\documentclass[acmtog, nonacm]{acmart}

\AtBeginDocument{%
  \fancypagestyle{plain}{%
    \fancyhf{} %
    \fancyfoot[L]{ACM SIGENERGY Energy Informatics Review}%
    \fancyfoot[R]{Volume 1 Issue 1, November 2021}}
  \providecommand\BibTeX{{%
    \normalfont B\kern-0.5em{\scshape i\kern-0.25em b}\kern-0.8em\TeX}}}
    
\let\oldmaketitle\maketitle
\renewcommand{\maketitle}{%
  \oldmaketitle%
  \thispagestyle{plain}%
  \pagestyle{plain}}

\usepackage[most]{tcolorbox}

\setcopyright{acmlicensed}
\copyrightyear{2018}
\acmYear{2018}
\acmDOI{XXXXXXX.XXXXXXX}

\acmConference[HotCarbon’25]{}{July 10-11th, 2025}{Cambridge, MA}
\acmISBN{978-1-4503-XXXX-X/18/06}

\usepackage{xurl}
\usepackage{subcaption}
\usepackage{xspace}
\usepackage{wrapfig}

\newenvironment{tightlist}{
\begin{list}{$\bullet$}{
    \setlength{\topsep}{.1em}
    \setlength{\partopsep}{0in}
    \setlength{\parskip}{0in}
    \setlength{\itemsep}{0in}
    \setlength{\parsep}{0in}
    \setlength{\leftmargin}{1em}
    \setlength{\rightmargin}{0in}
    \setlength{\itemindent}{0in}
}}
{\end{list}}

\newcommand{\tinyskip}{\vspace{3pt}}
\newcommand{\mypar}[1]{\tinyskip\noindent\textbf{#1.}\xspace}
\newcommand{\eg}{\text{e.g.,}\xspace}

\begin{document}

% \citestyle{acmauthoryear} % remove this line - references should be numbered 

%\title{Edge Centric Foundation Model Training: \\ Opportunities and Challenges}
\title{Towards Decentralized and Sustainable \\ Foundation Model Training with the Edge}


\author{Leyang Xue}
\email{leyang.xue@ed.ac.uk}
\affiliation{%
  \institution{The University of Edinburgh}
  \country{United Kingdom}
}

\author{Meghana Madhyastha}
\email{mmadhya1@jhu.edu}
\affiliation{%
  \institution{Johns Hopkins University}
  \country{United States}
}

\author{Randal Burns}
\email{randal@cs.jhu.edu}
\affiliation{%
  \institution{Johns Hopkins University}
  \country{United States}
}

\author{Myungjin Lee}
\email{myungjle@cisco.com}
\affiliation{%
  \institution{Cisco Research}
  \country{United States}
}

\author{Mahesh K. Marina}
\email{mahesh@ed.ac.uk}
\affiliation{%
  \institution{The University of Edinburgh}
  \country{United Kingdom}
}


\begin{abstract}
  Foundation models are at the forefront of AI research, appealing for their ability to learn from vast datasets and cater to diverse tasks. 
  Yet, their significant computational demands raise issues of environmental impact and the risk of centralized control in their development. 
  We put forward a vision towards decentralized and sustainable foundation model training that leverages the collective compute of sparingly used connected edge AI devices. 
  We present the rationale behind our vision, particularly in support of its sustainability benefit. 
  We further outline a set of challenges that need to be addressed to turn this vision into reality.
  %We propose leveraging distributed edge computing as a sustainable and decentralized approach to training these models, capitalizing on the energy efficiency and growing AI capabilities of edge devices along with their low utilization. 
  %\review{As the vast scale of edge device collectively provides the bandwidth and compute power that matched with cloud, we aim at a comparable training time.}
  %Achieving this vision presents several significant challenges, including: desiging and assessing edge-tailored and low carbon training methods; ensuring seamless and efficient training task orchestration across a diverse, dynamic landscape of edge devices; and holistic, dependable and efficient energy consumption tracking capability for edge devices. 
\end{abstract}



% \keywords{Do, Not, Us, This, Code, Put, the, Correct, Terms, for,



% \noindent\fbox{%
%     \parbox{\textwidth}{%
%       \small\textbf{Citation:} Xu, N., et al. \textit{Democratized and Sustainable Training and Inference of Foundation Models.} In Proc. of NSDI 2026.
%     }
%   }
% \vspace{1ex}
\maketitle

\begin{tcolorbox}[colback=yellow!10, colframe=black, boxrule=0.5pt,
  arc=2pt, left=4pt, right=4pt, top=4pt, bottom=4pt, enhanced,
  sharp corners, width=\columnwidth, boxsep=2pt]
\small
\textbf{Note:} This paper has been accepted to appear at \textit{HotCarbon 2025}.
\end{tcolorbox}
% \vspace{0.5em}  % optional spacing

\section{Introduction}
``Foundation models''~\cite{DBLP:journals/corr/abs-2108-07258}, which are models trained at scale on broad data that can then be adapted to a wide range of downstream tasks, are at the heart of the current AI revolution. 
Such foundation models leverage the power of generative AI and make AI a general-purpose technology, heralding it into the industrial age~\cite{ai-industry}.
They span domains as diverse as natural language processing (NLP)~\cite{gpt4}, computer vision (CV)~\cite{stable-diffusion}, software development~\cite{copilot}, networks~\cite{DBLP:conf/hotnets/LeSGS22,DBLP:conf/hotnets/Kotaru23}, biology~\cite{alphafold} and more. 

The immense promise that foundation models offer is in large measure owed to their scale; they exhibit an ``emergent'' behavior with model size and show a sharp rise in accuracy as they are scaled up beyond a point and trained with large amounts of data~\cite{scalelaw}.
This has resulted in the compute demand for their training skyrocketing in recent years, needing 1000s of AI accelerators (GPUs, TPUs, etc.)~\cite{ml-era}.
Such high compute resource requirements come with a significant economic cost (e.g., \cite{large-model-cost}) and environmental cost (\eg~\cite{DBLP:journals/corr/abs-2412-06288}), affordable to only a handful of global entities mostly those who run the cloud. This increasing centralization is a big impediment to the collective development of AI to benefit all~\cite{Bartoldson24,ai-review,borge2022deep}.
Not only that, there is an enormous environmental cost that is rising rapidly, again rooted in the compute-intensive nature of foundation model development~\cite{chatgpt-energy,DBLP:conf/mlsys/WuRGAAMCBHBGGOM22}.
Unsurprisingly then, AI has become one of the big four contributors to global ICT-related carbon footprint~\cite{knowles2021acm}.

In this paper, we put forward a vision centered on the ``edge'' as a distributed computing platform to drive future foundation model development in a decentralized and sustainable manner.
The key idea is to harness the spare compute across an amorphous collection of connected edge AI devices for foundation model training.
There are billions of such mostly idle edge devices across the globe that offer ample opportunity to this end. 
The approach we advocate builds on the pioneering works in the realm of volunteer computing~\cite{DBLP:journals/csur/MengistuC19}, aimed at harnessing the idle computing power from personal devices for distributed computing tasks. 
SETI@home~\cite{DBLP:journals/cacm/AndersonCKLW02} is a notable example that leveraged a million volunteered computers in the search for extraterrestrial intelligence.
Differently from these early works, our focus is on enabling decentralized and sustainable foundation model development via the edge at a lower overall carbon footprint, while maintaining accuracy as with cloud-based training. 
%constrained by requiring training speed to be comparable to that in the cloud.

The decentralized aspect of edge-based foundation model training is obvious. 
We make a case that it also enhances sustainability overall by presenting our rationale through a three-step argument, as outlined below and elaborated in \S\ref{sec:rationale}:

\begin{tightlist}    
    \item Edge devices are designed with energy efficiency in mind and increasingly feature AI accelerators to support ML tasks, including training. By comparing the energy consumption of cloud- and edge-based training through a series of experiments, we show that it can be several times more energy efficient to train with edge devices.
    
    \item Edge devices are mostly idle (75\% of the time in the case of smartphones) but have a high embodied carbon footprint. The latter is unrelated to their operational use and instead linked to manufacturing, supply chain, and recycling. So, utilizing them better helps amortize their high embodied carbon footprint.
    
    \item The above enables the opportunity to offload compute from the cloud to the edge, thereby helping reduce the cloud-side carbon footprint.
    This is because the baseline (embodied + operational) carbon footprint of edge devices would always be incurred simply by individuals owning them. 
    Their better use can yield net gains in carbon footprint due to their relatively better energy efficiency and ample idle time. 
    Our analysis shows that a 4–8× net carbon footprint reduction is possible by replacing the overall carbon footprint of a cloud GPU device with a small addition to the operational carbon footprint of a set of edge devices providing equivalent compute. 
\end{tightlist}

There have been some efforts in line with our vision aimed at decentralized ML training with volunteered devices~\cite{learning@home,dedloc,swarm} but they do not account for all the unique characteristics of the edge environment (device heterogeneity and dynamism) and crucially overlook the sustainability dimension altogether.
Significant challenges remain to be addressed to bring our vision to fruition, including:
(i)~\emph{Distributed training methods for edge}: Scalable and energy-efficient training requires edge-aware workload partitioning and communication minimization, as well as accounting for the carbon footprint;
(ii)~\emph{Training task orchestration at the edge}: Coordinating training tasks across dynamic edge environments demands fault tolerance and low-overhead, carbon-aware scheduling;
(iii)~\emph{Holistic, reliable and efficient energy consumption monitoring}: Measuring carbon impact at the edge requires accurate, low-overhead energy monitoring across all hardware components.
(iv)~\emph{User incentives, security and privacy}: 
These are key concerns that require maintaining a seamless user experience when their devices are leveraged for training, as well as efficient ways to achieve robustness and ensure training process isolation on user devices. 
%Preventing malicious participants and ensuring training process isolation on user devices requires low performance impact on training speed and robustness on training quality.

%Addressing these challenges will enable the technical capability to seamlessly and sustainably harness edge devices for foundation model development. 
%That can be the basis for exploring different approaches to incentivize user participation as well as other compelling use cases of an edge-centric distributed computing platform. 

\section{Background}

\subsection{Foundation Model Development Process}
\label{sec:model-dev-process}

\begin{figure}[t]
    \centering
    \includegraphics[width=\linewidth, page=3, trim={10cm 1.5cm 10cm 1cm},clip]{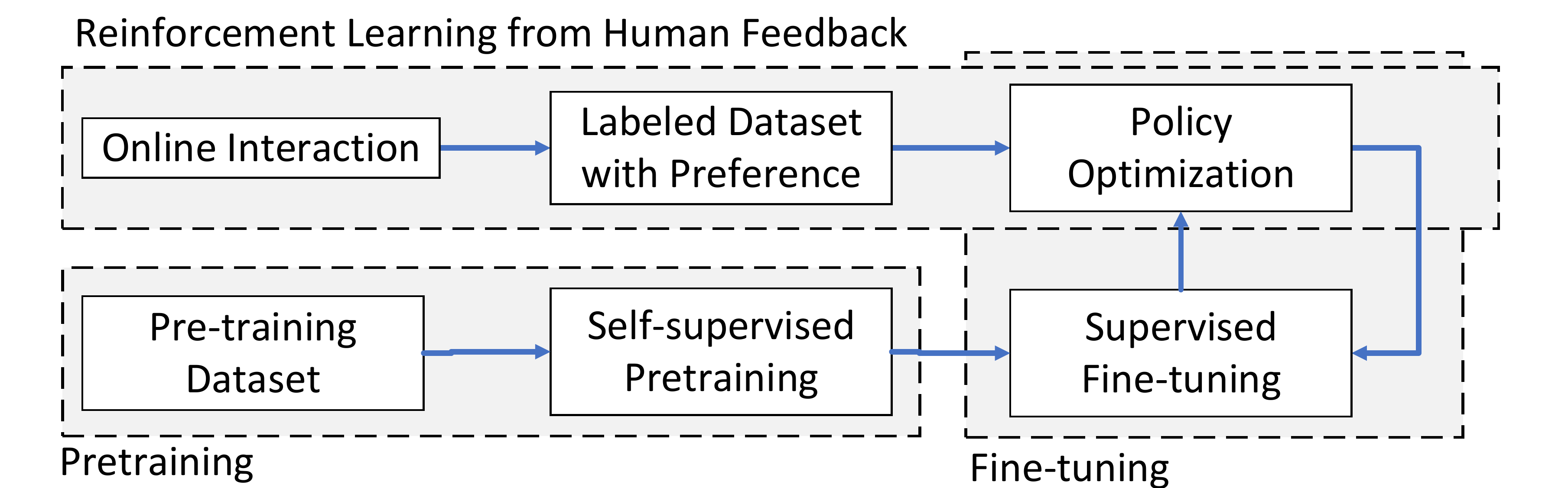}
    \vspace{-0.3in}
    \caption{Foundation model development process. }
    \label{fig:model-dev}
\end{figure}

Foundation models are characterized by three key traits: training on massive datasets, large model sizes, and broad generalization across tasks~\cite{DBLP:journals/corr/abs-2108-07258}. 
Large language models (LLMs), such as GPT-4, are a prominent example. Similar models exist for other modalities (e.g., Stable Diffusion~\cite{stable-diffusion} for vision) and for multimodal data (e.g., DeepSeek-VL2~\cite{deepseek-vl2}, QWen2.5-VL~\cite{qwen2.5-vl}).
Their development typically involves two phases (Fig.~\ref{fig:model-dev}): pre-training and fine-tuning.

% Foundation models have three key characteristics: trained on massive datasets, large model sizes and can be generalized to various tasks~\cite{DBLP:journals/corr/abs-2108-07258}. Large language models (LLMs) are a well-known example of foundation models that deal with language data. Foundation models also exist for other types of data (e.g., Stable Diffusion~\cite{stable-diffusion} for vision data) as well as multimodal foundation models that span multiple types of data (e.g., LLaVA-1.5~\cite{llava}). 
% The development of foundation models typically consists of two key phases: pre-training, and fine-tuning, as illustrated in Fig.~\ref{fig:model-dev}. 

\mypar{Pre-training}
In the pre-training phase, foundation models are trained on vast datasets (\eg 45TB in GPT-3)~\cite{gpt3} typically through self-supervised learning to develop a broad understanding of the world, capturing general patterns, structures, and knowledge from the data.
Pre-training requires all model parameters to participate as well as multiple levels of precision~\cite{zeropp}.
As the size of the dataset and model is often beyond a single node's capability, sharding of pre-training computation is required to scale it out across multiple computation and communication resources~\cite{alpa}.
Pre-training is not necessarily a one-time operation; recently, continual pre-training~\cite{continual-pretraining,ERNIE} has emerged to enable model pre-training over multiple rounds on domain-specific data. 

\mypar{Fine-tuning} Pre-training is followed by fine-tuning to specialize in specific tasks. 
Fine-tuning can be a lighter training that tunes models to specific benchmarks  (\eg 300GB data in FLAN~\cite{FLAN}) or capability of chat~\cite{llama2,mixtral}.
It can also be in the form of few-shot learning trained on only a few representative samples~\cite{gpt3}.
Fine-tuning not only has smaller data samples but also needs less computation when model parameters are partially fixed or quantized~\cite{qlora}. Post-training processes such as Reinforcement Learning from Human Feedback (RLHF)~\cite{DBLP:journals/corr/abs-2009-01325} also involve fine-tuning foundation models, actively with user/developer feedback. 
%RLHF also needs to create a reward model that has similar size to foundation models (\eg hundards of GB).
% RLHF process collects both user preference and developer preference to fine-tune the model more actively and frequently than pretraining.
RLHF is either a continual process that runs along with model inference or the model needs to be retrained in order to adapt to new data~\cite{zhang2024cppo}. 

Among the above two phases, pre-training is the most computationally heavy part followed by fine-tuning with RLHF. 
%The technique specified can be applied to all types of training.
% Unless otherwise specified, our focus is on pre-training in this work.

% \vspace{-0.2in}
\subsection{Components of Carbon Emission}\label{sec:carbon-source}

% Carbon emission

% \mypar{Components of carbon emission}
Carbon emissions for a computing device can be categorized into two main types: 
(i)~\emph{embodied carbon emissions} encompassing all the carbon released throughout the life cycle of the device before it starts to be used or after its disposal, including manufacturing, transportation, and installation of materials and products~\cite{carbonexplorer,patterson-phone-cloud,carbnon-metric};
(ii)~\emph{operational carbon emissions} arising due to electricity consumption during operation of the device for computation and communication (data movement between processors/accelerators and memory, over network), while accounting for power usage efficiency of the computing infrastructure and carbon intensity of the energy source powering the infrastructure~\cite{carbonexplorer,patterson-phone-cloud,carbontracker}.
% The proportion of embodied and operational carbon varies for different kind of devices, from mobile phones to data center accelerators.
$CO_2$ is one of seven different greenhouse gases (GHGs) that cause global warming, but it is the most common one.
So, the emissions due to different GHGs are expressed using $CO_2$ equivalent ($CO_2e$) as the common unit. 

Carbon accounting is used to track the emissions from organizations, sectors, etc. 
This is being done for machine learning (ML) workloads too. 
To account for ``operational'' carbon in the ML context, tools like MLCO$_2$~\cite{mlco2} and CarbonExplorer~\cite{carbonexplorer} combine the power consumed (\eg in $kW$), workload duration (\eg in hours ($h$)) and carbon intensity of the local electricity grid (\eg in $kgCO_2e/kWh$) to estimate the operational carbon footprint (\eg $kW \cdot h \cdot kgCO_2e/kWh$).
In contrast, the ``embodied'' carbon is harder to track as it involves several different aspects. 
Encouraged by the GHG protocol~\cite{ghg}, holistic reporting of carbon emissions for computing products is happening (\eg Apple~\cite{iphone15-env,macbook-16in-env}).

\section{Motivation}\label{sec:motivation}
\begin{figure}[t]
    \centering
    \begin{subfigure}{.48\linewidth}
        \includegraphics[width=\linewidth]{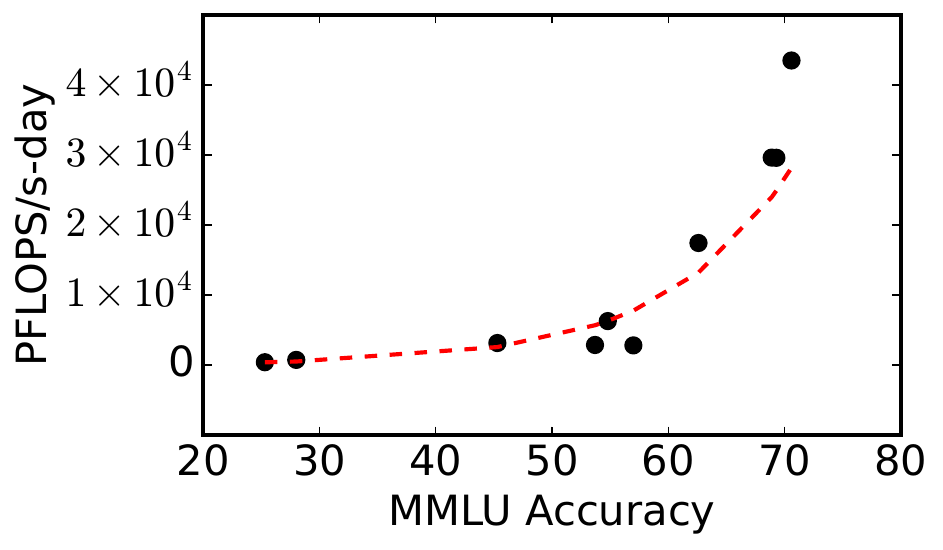}
        \vspace{-0.25in}
        \caption{Compute demand}
        \label{fig:compute-scaling}
    \end{subfigure}
    \begin{subfigure}{.48\linewidth}
        \includegraphics[width=\linewidth]{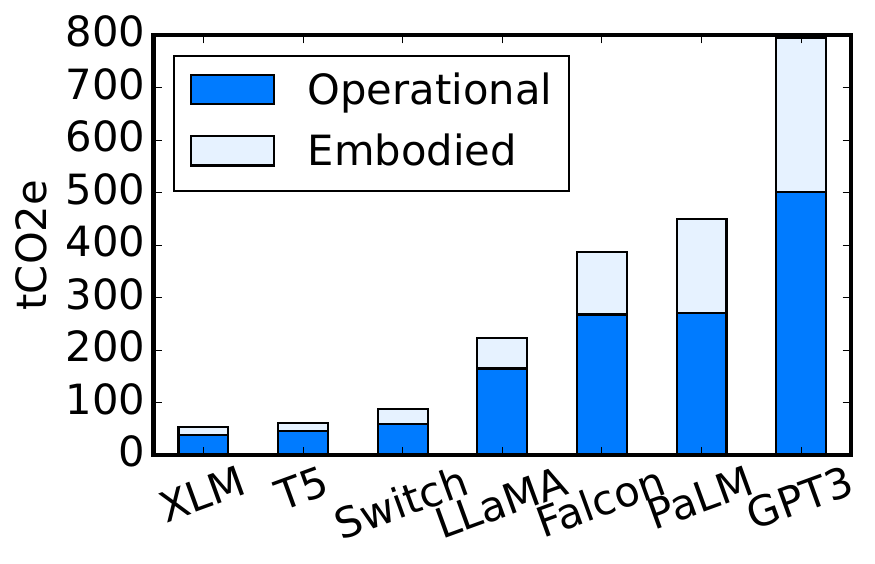}
        \vspace{-0.25in}
        \caption{Carbon footprint}
        \label{fig:carbon-scaling}
    \end{subfigure}
    \vspace{-0.15in}
    \caption{High costs of foundation model development.}
    % \vspace{-0.25in}
    \label{fig:model-problem}
\end{figure}

Compute requirements for model training in the deep learning era have exploded by 10-100 fold, increasing exponentially~\cite{ml-era}.
Considering the representative case of LLMs, the scaling laws~\cite{scalelaw} suggest that achieving linear growth in accuracy requires exponential growth in dataset size, model size and number of training iterations, that in turn require an exponential growth in compute demand (in terms of number and time on GPUs/TPUs). 
The aforementioned three ways of scaling have a multiplicative effect on the compute resource needed.
Fig.~\ref{fig:compute-scaling} illustrates this relationship between compute demand (in terms of petaflop-per-second needed to complete training in one day (PFLOP/s-day)~\cite{gpt3,plam}) and post-training model accuracy on the MMLU dataset~\cite{mmlu} for a range of models (shown in Fig.~\ref{fig:carbon-scaling}). 
Historically, models underwent size scaling (from XLM to GPT3), which has been shown to be essential to align with human performance~\cite{DBLP:journals/corr/abs-2303-18223}.
At present, models additionally leverage data scaling, given that LLMs in particular and foundation models generally are usually under-trained with insufficient data~\cite{DBLP:journals/corr/abs-2203-15556}.
Moreover, recent requirements on cross-domain~\cite{zhang2024cppo}, instruction-handling~\cite{llama2,mixtral} and multi-modal capability~\cite{fei2022towards,DBLP:conf/icml/DriessXSLCIWTVY23} of the models further contribute to the growth in data and training length. 

%\subsection{High Costs of Foundation Model Training}

This rapidly growing compute demand for large model training to advance the state-of-the-art has two undesirable implications: 

\mypar{1. Centralization} 
As model training has high compute demand, requiring tens of thousands of GPUs or other types of accelerators~\cite{openai-gpu}, it can be incredibly expensive. 
For instance, each training run of GPT-3 required at least \$5 million worth of GPUs~\cite{large-model-cost}. 
The overall cost of training such models is significantly more as many training runs are needed as they are developed and tuned. 
To make the situation even worse, essential techniques like neural architecture search might require the training of tens of thousands of neural networks~\cite{DBLP:conf/iclr/ZophL17}.
As a result, the computationally intensive nature of foundation model development and the associated prohibitively high economic costs naturally favor few resource-rich organizations, leading to increased centralization of the ecosystem~\cite{Bartoldson24,ai-review,borge2022deep}. 

\mypar{2. Unsustainable} The data center infrastructure to train LLMs and foundation models consumes an enormous amount of power with a potentially large carbon footprint. 
For instance, training ChatGPT has a 10 Gigawatt-hour (GWh) energy consumption, equivalent to the yearly consumption of 1000 US households~\cite{chatgpt-energy}.  
To drive home this point, in Fig.~\ref{fig:carbon-scaling}, we report the estimated carbon footprint in tons of $CO_2e$ ($tCO_2e$) for a range of foundation models to achieve the increasing levels of accuracy on the MMLU dataset, as in Fig.~\ref{fig:compute-scaling}.
For this result, we use the carbon emission data from the papers of the models~\cite{t5,llama2,gpt4,gropher,plam} where available; Otherwise, we use LLMCarbon~\cite{faiz2024llmcarbon}.
This shows that the accuracy advancement comes at the expense of exponential growth in carbon footprint. 
Both the embodied carbon footprint (linked to renewing GPUs every 3-4 years~\cite{carboncurtain}) as well as the operational carbon footprint of training in a data center~\cite{patterson-cloud,carbon-aware-cloud} contribute to this trend. 
Crucially, this path of advancement in foundation model is unsustainable~\cite{DBLP:conf/mlsys/WuRGAAMCBHBGGOM22}. 

\section{Edge Centric Foundation Model Training}

\subsection{Vision}\label{sec:vision}

We posit that the edge can offer an effective alternative for decentralized and sustainable foundation model development. 
The key idea is to harness the spare (unused) compute across an amorphous collection of edge devices for foundation model training.
%while make it comparable to cloud based training in terms of training time.
Crucially, we seek to lower the overall carbon footprint associated with foundation model training via the edge while maintaining the accuracy as with cloud based training.
%Crucially, we seek to lower the overall carbon footprint associated with foundation model training with the edge while constraining to match training time with cloud based training.
Just as the notion of edge computing itself is broad, there are a wide variety of edge devices~\cite{edge-dl-survey}. 
For our context, we focus on edge devices that are network-connected and equipped with AI accelerators.
Examples include smartphones and laptops with GPUs and ML accelerators (e.g., Google Edge TPU, Apple M3/A16, Samsung NPU).
%We do not consider IoT devices that are limited in their computational and memory capabilities. 
There exist billions of such suitable edge devices. 
Just considering smartphones alone, there are currently over 6.5 billion of them~\cite{patterson-phone-cloud}. 
Moreover, they are mostly idle; typically usage ranges from 3-6 hours a day~\cite{phone-usage}.
With the right incentives, the massive scale and global scope of such edge devices offers plentiful opportunity to select a subset of them with collective compute capability matching that in the cloud to engage in the process of foundation model training. 

%\mypar{Inspirations}
The approach we advocate to leverage idle periods of edge devices is inspired by prior efforts like SETI@home~\cite{DBLP:journals/cacm/AndersonCKLW02}, Computing While Charging~\cite{DBLP:conf/conext/ArslanSSMSK12} and more generally, volunteer computing~\cite{DBLP:journals/csur/MengistuC19}.
The general idea behind these works is that performance-insensitive yet computationally heavy tasks can be naturally distributed among personal devices to reduce the cost.
The BOINC framework~\cite{DBLP:journals/grid/Anderson20} realized this concept, tapping into approximately one million computers volunteered by 600,000 individuals to power fundamental scientific research, especially search for extraterrestrial intelligence~\cite{DBLP:conf/sc/AndersonCA06,korpela2001seti}.
Subsequently, Arslan et al.~\cite{DBLP:conf/conext/ArslanSSMSK12} explored the potential for exploiting unused compute in mobile devices during charging.
Our work builds on these above pioneering works with the vision of turning the edge into a seamless distributed computing platform, motivated by foundation model training as a compelling and timely use case.

% \vspace{-0.15in}
\subsection{Rationale}\label{sec:rationale}

\mypar{Energy-Efficient Edge Devices with AI Capabilities}
\label{edge-device-energy-efficiency}
Energy efficiency is a principal consideration when designing edge device hardware due to the following factors:
(i)~\emph{Power constraints:} Edge devices typically are battery-powered, so energy-efficient design is necessary to extend the device's operational lifetime between charges~\cite{apple-m2}.
%Even when the devices are charging, computation tasks on the device can still benefit from the energy efficiency due to the maximum power cap. 
% The ARM Big.LITTLE architecture makes these edge devices energy efficient through the coupling of slower but battery efficient cores (LITTLE) with more powerful but power hungry cores(Big) in a multicore system. 
(ii)~\emph{Thermal constraints:} Unlike in cloud data centers, these devices typically lack sophisticated cooling systems. Thus, they must be designed to minimize heat production and operate effectively within their thermal budgets~\cite{DBLP:conf/micro/ShaoCVZFJKKPRTZ19}.
%In the sense of energy efficiency, passive cooling uses all power for computing rather than sharing that with a cooling system.
(iii)~\emph{Task specialization:} Modern edge devices are increasingly well-suited for machine learning workloads, as they feature specialized computation units for matrix multiplication, low-precision operations, etc. 
Hardware heterogeneity is a hallmark of edge devices, both across different devices and within a single device (e.g., the ARM Big.LITTLE architecture). 
These designs maximize both task performance and power efficiency~\cite{DBLP:conf/iccad/KrishnanGMWCSO022,DBLP:journals/dt/KrishnanMCSOC20}.

\begin{figure*}[t]
    \centering
    \begin{minipage}{.33\linewidth}
        \setlength\tabcolsep{2.5pt}
        \begin{tabular}{|c|c|c|c|}
        \hline
           Device  & Power &  Time & Energy\\
        \hline
           Smartphone & 10W & 3510s  & 9.75Wh  \\
           (Snapdragon 888)&&&\\
           \hline
     %      Jetson Orin  & 3600 & 12 & 12 \\
           Laptop   & 15W & 480s & 2Wh\\
            (M2 Pro)&&&\\
            \hline
            Cloud GPU & 220W &    250s & 15.28Wh \\
              (NVIDIA A5000) &&&\\
        \hline
        \end{tabular}
        % \vspace{-0.15in}
        \captionof{table}{Energy efficiency of ML training with single device considering OPT-125m model.}
        \label{tab:training-energy}
    \end{minipage}
    \hfill
    \begin{minipage}{.25\linewidth}
        \centering
        \setlength\tabcolsep{2.5pt}
        \begin{tabular}{|c|c|}
        \hline
           Device  & Energy\\
        \hline
            Cloud GPU & 152Wh\\
              (NVIDIA A5000) &\\
           \hline
           4 Laptops   & 27Wh\\
            (M2 Pro)&\\
        \hline      
        15 Smartphones & 98Wh\\ 
           (Snapdragon 888)&\\
            \hline
        \end{tabular}
        % \vspace{-0.15in}
        \captionof{table}{Energy efficiency of distributed ML training across edge devices with DT-FM and OPT-1.3B model.}
        \label{tab:training-energy-13b}
    \end{minipage}
    \hfill
    \begin{minipage}{.40\linewidth}
        \centering
        \includegraphics[width=\linewidth]{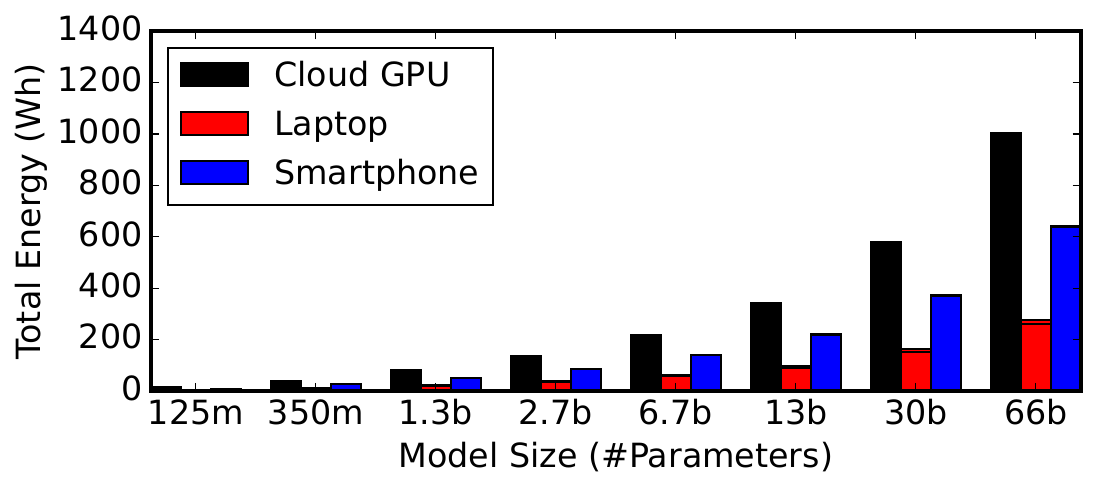}
        \vspace{-0.35in}
        \caption{Training energy consumption comparison across cloud and edge settings for different OPT model sizes with an idealized distributed training method.}
        \label{fig:model-size}
    \end{minipage}
    \vspace{-0.2in}
\end{figure*}

To demonstrate the energy efficiency and suitability of edge devices for ML training, we conduct experiments with three different devices: (1) Smartphone with Snapdragon 888 SoC; (2) Apple MacBook Pro laptop with M2 Pro chip; (3) Cloud GPU represented by NVDIA A5000. 
We consider models from the OPT series~\cite{opt}.
First, we pick a small OPT-125m model that can be trained using any of the above three devices. 
Table~\ref{tab:training-energy} shows the result of training this model on the MMLU dataset~\cite{mmlu} for 100 steps with a batch size of 16 and sequence length 512. 
We observe that, although edge devices fare worse by 2-10x in terms of training time compared to the cloud GPU case, they have 1.5-7.5x lower energy consumption due to 15-20x lower power consumption. 

Next, we choose a larger OPT-1.3B whose training can still be done with a {\em single} cloud GPU but requires multiple edge class devices. 
Specifically, assuming a homogeneous set of edge devices, this model needs 4 laptops and 15 smartphones to hold all parameters and training states.
For distributed edge training, we use the state of the art (SOTA) DT-FM method~\cite{dtfm} that employs a combination of data and pipeline parallelism. 
For this experiment, we keep the dataset and training hyperparameters the same as in the last experiment.
Assuming homogeneous and symmetric network bandwidth of 10MB/s for each edge device and 0.5W peak power for their WiFi communication modules~\cite{junkyard},
results in Table~\ref{tab:training-energy-13b} show that distributed training across edge devices still offers 1.5-5x better energy efficiency compared to training with a cloud GPU, even after accounting for communication related energy consumption.

%(ii) the network communication has bandwidth of 10MB/s through broadband with 0.5W peak power on the wireless communication module~\cite{junkyard}.

Building on the above two experiments where two specific model sizes are considered, we now study generalization across different model sizes. 
To this end, for fair comparison between cloud and edge settings, we consider an {\em idealized}\footnote{The idealized training method models training as a series of operators in a directed acyclic graph (e.g., \cite{alpa}), with a controller distributing the computation of operations across devices. The communication load includes the model size and all intermediate results. The controller can aggregate gradients locally during forward and backward propagation without additional communication, as devices transmit output from each operator back to the controller without peer-to-peer broadcasting.
% This ideal training method distributes training computation across devices via data and pipeline parallelism.
In addition with this ideal method, the gradient for each parameter is transmitted only once and the intermediate result in each layer is transmitted from devices only once so that the total data transmitted is model size + (intermediate size * number of layers) for each batch.
%Moreover, all transmissions can utilize the aggregated bandwidth across all devices. 
} training method in order to factor out the differences arising due to specific distributed training methods employed in the cloud (e.g., \cite{megatron}) and those designed for the edge (e.g., \cite{dtfm}).
Fig.~\ref{fig:model-size} shows the energy consumption comparison for distributed training with cloud and edge devices as a function of varying model size, again considering OPT series of models.
The idealized distributed training method as outlined above is used throughout the paper.
Note that depending on the model size, multiple devices are required for training in both cloud and edge settings. 
From the results in Fig.~\ref{fig:model-size}, we observe that the energy efficiency of edge devices relative to cloud GPU devices contributes to lower overall training related energy consumption with edge devices. 
This is particularly pronounced with the laptop case. 
In general, these results highlight the potential for lowering training related energy consumption with edge devices by 1.5-4x compared to the cloud case across a range of model sizes.

\begin{figure}[t]
    \centering
    \begin{minipage}{.48\linewidth}
        \centering
        \includegraphics[width=\linewidth]{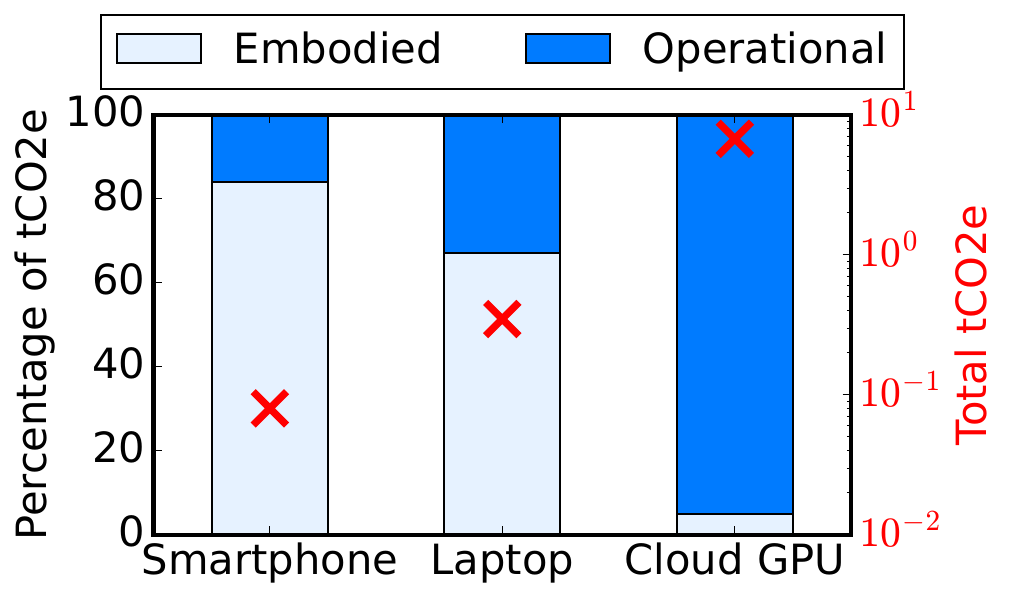}
        \vspace{-0.25in}
        \caption{Single device carbon footprint comparison: percentage breakdown between embodied and operational (left y-axis) and total absolute carbon footprint (right y-axis).}
        \label{fig:carbon-components}
    \end{minipage}
    \hfill
    \begin{minipage}{.48\linewidth}
        \centering
        \includegraphics[width=\linewidth]{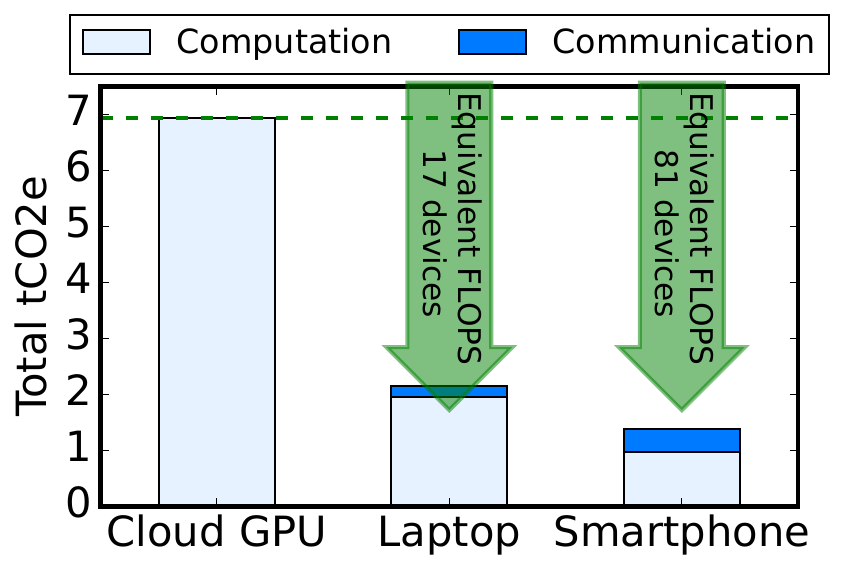}
        \vspace{-0.25in}
        \caption{Carbon emission over 3 years. Total carbon footprint on edge devices is lower when computation capacity is matched with cloud.}
        \label{fig:training-energy-hardware}
    \end{minipage}
\end{figure}

% \begin{figure*}[t]
%     \centering
%     \begin{minipage}{.28\linewidth}
%         \centering
%         \includegraphics[width=\linewidth]{figures/device_carbon.pdf}
%         \vspace{-0.25in}
%         \caption{Single device carbon footprint comparison: percentage breakdown between embodied and operational (left y-axis) and total absolute carbon footprint (right y-axis).}
%         \label{fig:carbon-components}
%     \end{minipage}
%     \hfill
%     \begin{minipage}{.29\linewidth}
%         \centering
%         \includegraphics[width=\linewidth]{figures/shift_carbon.pdf}
%         \vspace{-0.25in}
%         \caption{Carbon emission over 3 years.}
%         \label{fig:training-energy-hardware}
%     \end{minipage}
%     \begin{minipage}{.40\linewidth}
%         \centering
%         \includegraphics[width=\linewidth]{figures/comm_overhead.pdf}
%         \vspace{-0.25in}
%         \caption{Comparison of SOTA DT-FM with idealized training method to illustrate limitation of existing distributed ML training methods for edge. Hatch shows computation cost (bottom) and the rest shows communication cost (top).}
%         \label{fig:communication-energy}
%     \end{minipage}
% \end{figure*}

% \begin{figure}[t]
%     \centering
%     \includegraphics[width=.8\linewidth]{figures/device_carbon.pdf}
%     \vspace{-0.2in}
%     \caption{Single device carbon footprint comparison: percentage breakdown between embodied and operational (left y-axis) and total absolute carbon footprint (right y-axis).}
%     \label{fig:carbon-components}
% \end{figure}

\mypar{Low Utilization and High Embodied Carbon Footprint}\label{util-embodied}
As already noted above, edge devices often stay idle. For smartphones, this is at least 75\% of the time~\cite{phone-usage,mobileminer,DBLP:conf/conext/ArslanSSMSK12}. 
At the same time, these devices possess powerful compute capabilities to efficiently perform AI tasks including training. 
On the flip side, edge devices have a rather high embodied carbon footprint, as highlighted in Fig.~\ref{fig:carbon-components} that compares the carbon footprint of mobile and laptop devices against a data center GPU device (H100) over a 3-year period. 
For the mobile device, we use the average reported data from Apple's official product environmental report of iPhone 15~\cite{iphone15-env}.
For the laptop case, we use the data from a similar report for MacBook Pro for its 3-year lifetime~\cite{macbook-16in-env}.
We use NVIDIA H100 GPU as a representative data center GPU device and estimate its carbon footprint following the methodology from MLCO$_2$~\cite{mlco2} and using the carbon intensity of the electricity grid averaged across North America and Europe.
The embodied carbon per GPU is estimated as one eighth of the server footprint~\cite{patterson-phone-cloud}, given a typical GPU server has 8 GPUs. 

% \begin{wrapfigure}[13]{rI}{0.5\linewidth}
%     \vspace{-12pt}
%     \centering
%     \includegraphics[width=\linewidth]{figures/device_carbon.pdf}
%     \vspace{-0.3in}
%     \caption{Single device carbon footprint comparison: percentage breakdown between embodied and operational (left y-axis) and total absolute carbon footprint (right y-axis).}
%     \label{fig:carbon-components}
%     \vspace{-12pt}
% \end{wrapfigure}

We make two key observations from this comparison: (1) carbon footprint for edge devices is dominated by embodied carbon footprint, over 80\% for mobile devices (see left y-axis), in line with the observation in previous studies~\cite{patterson-phone-cloud}.
Low utilization of edge devices contributes to amplifying the proportion of their embodied carbon footprint.
On the other hand, operational carbon footprint is significant for data center GPU devices. 
(2) Data center GPU devices have significantly higher absolute amount of carbon footprint for the compute capability they offer (see right y-axis). 
For example, compared to the laptop case, data center GPU has at least an order of magnitude higher carbon footprint for 5x more compute capability (H100 has 267 TFLOPS versus 53 TFLOPS of M2-Ultra for FP16 computation). 
Considering the cloud data center infrastructure as a whole, \eg further including footprint for CPUs and server cooling, make this comparison worse for the data center case.
% Considering the cloud data center infrastructure as a whole would make this comparison worse for the data center case. 

% \vspace{-0.15in}
\mypar{Cloud to Edge Computation Offloading Opportunity}\label{offloading-opportunity}
We further observe that the carbon footprint associated with personal devices at the edge (as shown in Fig.~\ref{fig:carbon-components}) would always be incurred simply by their ownership and baseline use. 
This includes the rather high embodied carbon footprint plus the operational carbon linked to the typical use of an edge device. 
Better utilizing them (with a corresponding marginal increase in their operational carbon footprint) can help amortize their high embodied carbon footprint. 
Crucially, doing so has a bigger payoff in enabling the computational tasks such as foundation model training to be shifted/offloaded from the cloud to the edge, thereby reducing the corresponding carbon footprint (both embodied and operational) on the cloud side. 

Fig.~\ref{fig:training-energy-hardware} highlights the aforementioned offloading opportunity, reusing the data from Fig.~\ref{fig:carbon-components}.
Considering a 3-year replacement cycle across the board, the total carbon footprint for the data center GPU (H100) is estimated to be 7 $tCO_2e$ following the methodology from MLCO$_2$~\cite{mlco2} and using average carbon intensity for Europe and North America for each of the years in the period 2021-23~\cite{carbonfootprint}. 
To obtain the same compute capability in terms of FLOPS, we estimate needing 15 (M2 Pro) laptops or 69 (Snapdragon 888) smartphones assuming an additional 8 hour daily usage per device while charging~\cite{DBLP:conf/conext/ArslanSSMSK12,patterson-phone-cloud,apple-m2}. 
Increase in operational energy use and corresponding carbon footprint for both types of edge devices is shown in Fig.~\ref{fig:training-energy-hardware}. 
We observe that the computation from a data center GPU device can be fully offloaded to smartphone (laptop) devices with equivalent compute, resulting in a net reduction of 8x (4x) in total carbon footprint. 

% \begin{wrapfigure}[10]{rI}{0.5\linewidth}
%     \vspace{-12pt}
%     \centering
%     \includegraphics[width=\linewidth]{figures/shift_carbon.pdf}
%     \vspace{-0.3in}
%     \caption{Carbon emission over 3 years.}
%     \label{fig:training-energy-hardware}
%     \vspace{-10pt}
% \end{wrapfigure}

% \begin{figure}
%     \centering
%     \includegraphics[width=.8\linewidth]{figures/shift_carbon.pdf}
%     \vspace{-0.2in}
%     \caption{Carbon emission over 3 years.}
%     \label{fig:training-energy-hardware}
% \end{figure}

Fig.~\ref{fig:training-energy-hardware} also shows the increased communication related operational carbon footprint when edge devices participate in foundation model training 8 hours daily while they are being charged.  
Here we consider training the OPT-1.3B.
We estimate this communication related carbon footprint using the WiFi communication related carbon emissions from prior work~\cite{junkyard} for single devices and multiply that with the equivalent number of edge devices (15 laptops or 69 smartphones to match the cloud GPU's compute capability).
We only consider a single cloud GPU in isolation for this analysis, and thus the communication overhead in that case is zero.
We observe that even after accounting communication related carbon footprint, offloading to smartphone (laptop) type edge devices can still result in net reduction of 6x (3.5x) in total carbon footprint.

% \vspace{-0.15in}
\subsection{Related Work}\label{sec:current-effort}
% \subsection{Faster Training is Not Always Better}\label{sec:current-limitation}

% \todo{GPU-hr maps to energy, use hr}
Foundation model training is computationally intensive, requiring a large number of accelerators~\cite{openai-gpu} and so is expensive.
%As previously noted in \S\ref{sec:motivation}, this is not only expensive limiting access to a few organizations who can afford the enormous compute resources in the cloud~\cite{large-model-cost} but also has a large environmental cost.
To cut the costs, recent works use spot instances (e.g., \cite{bamboo,doll,dtfm}) or multiple different clouds~\cite{skypilot} but these do not lower the carbon footprint associated with cloud computing facilities.

%%%
%Recently, there have been some algorithmic oriented proposals advocating decentralized machine learning training in the sense of volunteer computing~\cite{learning@home,dedloc,swarm}.
%In contrast, we take a systems oriented view. 
%Crucially, as discussed later in \S\ref{sec:current-effort}, these prior works (e.g., \cite{dedloc}) do not account for all the unique characteristics of the edge environment and do not at all consider the sustainability dimension.

In line with our vision in \S\ref{sec:vision}, some recent works pursued decentralized training with volunteered devices~\cite{learning@home,dedloc,swarm}. 
However, these works do not fully consider the unique characteristics of the edge environment that include device heterogeneity (in terms of compute and communication) as well as dynamism (in terms of participation and failures). 
Some of these works are also not suitable for large model training (e.g., \cite{dedloc}).
Crucially, all these prior efforts aim solely at training efficiency and overlook the sustainability dimension altogether.
Focusing only on optimizing training performance without regard to sustainability may result in an undesirably high carbon footprint. 
For example, devices with a higher embodied carbon footprint can be excessively used to maximize the training speed. 
As another example, devices in regions powered by the grid with high carbon intensity can end up being used, causing a higher operational carbon footprint as suggested by the analysis in \cite{patterson-phone-cloud}. 

Federated Learning (FL)~\cite{DBLP:conf/aistats/McMahanMRHA17} is a well-known distributed learning approach designed with edge devices in mind. 
However, the scenario targeted by FL is different from our aim to leverage edge devices for foundation model training in terms of the following three aspects: 
(i)~\emph{Model}: FL is akin to data parallelism~\cite{DBLP:conf/osdi/LiAPSAJLSS14} and typically assumes that the model can fit within the memory of a single device~\cite{totoro}. In contrast, foundation models including LLMs exceed the memory capacity of individual devices and so need to be sharded across devices. 
(ii)~\emph{Data}: FL is designed to ensure the privacy of data held on individual devices~\cite{DBLP:conf/aistats/McMahanMRHA17}, whereas foundation models (e.g., \cite{llama2}) are typically trained on large publicly available datasets, so our setting is more flexible with respect to data movement. 
%Non-iid data is natural for FL as data cannot be shifted outside user device. While foundation model relax the restriction. When the data is iid, FL falls back to data parallel.
% FL aims to resolve non-iid distribution of data across devices, which is naturally non-iid. When the data is iid, this falls back to data parallel.
(iii)~\emph{Algorithm}: Non-IID data is a defining characteristic of the setting that FL targets~\cite{DBLP:conf/aistats/McMahanMRHA17} which necessitates the use of specialized optimizers and client selection for higher model accuracy~\cite{DBLP:conf/aistats/McMahanMRHA17,feddist,fedadam}. Our foundation model training setting does not present such a requirement.
Moreover, even for training models that can fit in a single edge device, a recent analysis~\cite{patterson-phone-cloud} highlighted energy efficiency concerns with the training method underlying FL. 
%In particular, it was shown that FL based training using smartphones can consume 12x more energy compared to cloud based training with two-thirds of that energy consumption due to the FL server. 
%Designing efficient distributed foundation model training methods tailored for the edge environment characteristics is in fact a challenge, as we elaborate in \S\ref{sec:challenge}.
%The analysis in \cite{patterson-phone-cloud} also suggests that naively choosing smartphones that either come with poor charging efficiency or are charged using carbon intensive energy source can significantly increase the carbon footprint of smartphone based ML model training. 
%So it is important to carefully select edge devices when orchestrating foundation model training at the edge, a challenge we elaborate further in \S\ref{sec:challenge}.

Recently, \cite{junkyard} proposes repurposing {\em discarded} phones to create a compute cluster they call ``Junkyard Computer''. 
Similarly, accelerators from discarded phones can be put together as a server~\cite{DBLP:conf/hotcloud/ShahradW17,DBLP:conf/usenix/ZhangFSLLY00WX24}.
The motivation is to effectively extend the lifetime of such discarded phones and amortize their high manufacturing related embodied carbon footprint.
At a broad level, we share the same motivation.
However, our means to that end are different and complementary via better utilizing edge devices that are operational and without modification.
Furthermore, unlike \cite{junkyard}, we have a particular focus on challenges associated with large ML model training with edge devices. 
For this purpose, operational devices have higher energy efficiency and better capability for training (e.g., with cutting-edge edge device accelerators).

\section{Challenges and Potential Solutions}
\label{sec:challenge}

Decentralized training of foundation models using edge devices promises a democratized and sustainable path to AI. However, realizing this vision requires overcoming several key challenges. Unlike centralized cloud infrastructures, edge environments are inherently heterogeneous, intermittently available, and constrained in compute, memory, and energy capacity. To make edge-based training viable and competitive with cloud-based alternatives, it must achieve comparable training throughput and time-to-convergence, and maintain model fidelity using the same architectures, optimizers, and hyperparameters.

\mypar{Distributed training methods for the edge}
Training foundation models at the edge must account for heterogeneous compute, variable connectivity, and dynamic device participation, all while minimizing carbon footprint. Although edge devices collectively offer ample compute, communication overhead grows rapidly with scale and can potentially dominate energy usage. 
Existing compression techniques~\cite{zeropp,DBLP:journals/corr/abs-2206-01299,gptq,smoothquant,qlora} reduce communication but are typically limited to fine-tuning due to accuracy concerns. 
While energy costs of wide-area data transmission are relatively lower—around 0.001 kWh/GB~\cite{caribou,evergreen}—compared to local computation (at least 0.02 kWh/GB), the overhead still compounds with inefficient communication patterns and should not be overlooked.
The key challenge is to design distributed training methods that optimize both speed and carbon footprint by flattening communication-related energy costs as the system scales.
A potential solution is to use hybrid forms of parallelism strategies (e.g., data + tensor parallelism) in conjunction with dynamic workload scheduling that maximizes parallelism and adapts to diverse characteristics across devices including bandwidth availability.

% \review{
% We begin by considering the energy consumption associated with ISP networks and data transmission. Since ISP infrastructure is typically operated at full power regardless of load to meet performance requirements~\cite{rollin2024sleep}, additional communication does not increase energy use in these components. Moreover, the energy cost of data transmission over the WAN is approximately 0.001~kWh/GB~\cite{caribou,evergreen}, which is negligible compared to the per-GB energy cost of computation on any device (at least 0.02~kWh/GB).
% }

\mypar{Training task orchestration at the edge}
Orchestrating training across edge devices must balance fault tolerance, training speed, and carbon footprint for effective operation within a highly dynamic and resource-constrained environment. 
A unique aspect of edge devices is that they are susceptible to thermal throttling. This means sustained compute loads cause an increase in device temperature and trigger hardware-imposed slowdowns, which in turn increase latency and reduce energy efficiency.

On the fault tolerance front, traditional methods (\eg checkpointing~\cite{doll,mario,tenplex}, replication~\cite{bamboo}, and recomputation~\cite{swarm,asteroid}) pose trade-offs between carbon footprint and recovery latency, with replication increasing carbon costs and recomputation risking slowdowns. 
Achieving seamless fault tolerance with minimal overhead requires identifying Pareto-optimal strategies suited for dense, communication-bound edge training workloads. 
Furthermore, sustainable orchestration must incorporate thermal- and carbon-aware device selection and remain lightweight compared to existing frameworks~\cite{mlcube,kubeedge,DBLP:journals/corr/abs-2207-01577}.

In addition, efficient management of millions of edge devices and connections using a limited number of server-side resources requires highly scalable and low-overhead coordination mechanisms.
A promising direction is to leverage historical device activity and energy profiles to guide carbon-efficient scheduling, while enabling lightweight, decentralized orchestration with built-in support for preemptible execution and fast state recovery via proactive partial replication or reactive live migration.
Techniques such as hierarchical orchestration architecture, event-driven communication, sparse state synchronization, and pub-sub messaging systems can further minimize coordination overhead.

\mypar{Holistic, reliable and efficient energy consumption monitoring}
Accurate energy monitoring is essential for environmentally sustainable training on edge devices, as it enables quantifying the operational carbon footprint, identifying inefficiencies, and informing carbon-aware scheduling and incentives. Reliable measurements are essential to evaluate and optimize the environmental impact of training workloads, particularly in heterogeneous and resource-constrained edge environments. Existing methods are often GPU- or cloud-focused~\cite{greenscaler,irene,faiz2024llmcarbon,carbonexplorer,mlco2} and fail to account holistically for other system components like memory, storage, and networking. Moreover, software-based tools offer only coarse-grained measurements~\cite{nvml,intel-rapl}, missing the sub-millisecond energy dynamics of ML operations and risking misattribution of energy use~\cite{carbontracker}. The key challenge is to create cross-platform solutions that remain lightweight, accurate, and holistic across diverse edge devices.
A promising direction is to develop accurate, component-level energy models that can infer fine-grained consumption patterns and use coarse-grained measurements for periodic calibration.

% Accurate energy monitoring is essential for environmentally sustainable training on edge devices, but current tools fall short in both scope and precision. Existing methods are often GPU- or cloud-focused~\cite{greenscaler,irene,faiz2024llmcarbon,carbonexplorer,mlco2} and fail to account holistically for other system components like memory, storage, and networking. Moreover, software-based tools offer only coarse-grained measurements~\cite{nvml,intel-rapl}, missing the sub-millisecond energy dynamics of ML operations and risking misattribution of energy use~\cite{carbontracker}. The key challenge is to develop cross-platform, fine-grained energy monitoring methods that remain accurate and holistic.

\mypar{Security and privacy}
Security and privacy are critical to enable decentralized and sustainable distributed training, as trust in the platform underpins long-term, collaborative model development. 
Attacks on data or model parameters can waste compute resources and increase carbon cost through retraining or recovery. 
While current solutions like confidential computing~\cite{intelconf,nvdaconf} or encrypted training~\cite{DBLP:journals/csur/WoodNK20,DBLP:conf/nips/ZhangZSCL24} focus on local and computation-rich settings, scalable and low-overhead protections are needed for distributed, resource-constrained environments. 
In addition, these methods do not prevent data poisoning attacks such as adversarial model updates.
Although the use of public datasets and open-source models can mitigate some privacy concerns, the training platform must still operate under zero-trust assumptions, as it may run alongside other user applications and thus be vulnerable to attacks.
A potential solution is to incorporate lightweight, decentralized anomaly detection and attestation protocols that verify input data integrity and model behavior in real-time, enabling secure training without incurring the overhead of heavy cryptographic methods.

\mypar{User incentives}
To build a sustainable large-scale training platform leveraging user devices, incentivizing participation must go beyond economic rewards to include environmental awareness. 
Current systems offer service credits (e.g., NetMind~\cite{netmind}, Aioz~\cite{aioz}), but few encourage behaviors like charging during cleaner energy hours or using efficient chargers. 
Since training adds compute and memory load, it can degrade user experience and charging efficiency, necessitating careful coordination. 
The challenge lies in enabling preemptible, seamless training with fault-tolerant recovery, informed by user activity and energy conditions.
It would be more beneficial to reward users not only for contributing compute but also for aligning their participation with low-carbon and high-efficiency energy windows (\eg more hours during days with solar power), using lightweight client-side monitors to track availability, charging state, and responsiveness.

\section*{Acknowledgments}

We thank the anonymous reviewers for their helpful comments and suggestions that greatly improved this paper. 
The authors would like to also specially thank Jon Crowcroft for his review and helpful feedback on the paper. 

%\newpage

\bibliographystyle{ACM-Reference-Format}
\bibliography{sample-base}

\appendix

\end{document}